%% file: main.tex
\documentclass[11pt,a4paper]{article}
\usepackage[hyperref]{naacl2021}
\usepackage{times}
\usepackage{latexsym}
\usepackage{kky}
\usepackage{xspace}

\usepackage[T1]{fontenc}

\usepackage[utf8]{inputenc}

\usepackage{microtype}
\usepackage{booktabs}
\usepackage{subcaption}

\newcommand{\xmodel}{{\scshape amber}\xspace}

\usepackage{mathtools}
\usepackage{environ}
\NewEnviron{resizealign}{\sbox0{
    $\begin{matrix}\displaystyle\BODY\end{matrix}$}%
  \sbox1{$(\theequation)$}%
  \sbox2{\parbox{\dimexpr \wd0 + 1\wd1}%
    {\begin{align}\BODY\end{align}}}
  \noindent\resizebox{\columnwidth}{!}{\usebox2}%
}

\title{Explicit Alignment Objectives for Multilingual Bidirectional Encoders}

\author{Junjie Hu$^1$\thanks{*Work partially done at Google Research.}, Melvin Johnson$^2$, Orhan Firat$^2$, Aditya Siddhant$^2$, Graham Neubig$^1$ \\
  $^1$Carnegie Mellon University, $^2$Google Research \\
  \texttt{\{junjieh,gneubig\}@cs.cmu.edu,\{melvinp,orhanf,adisid\}@google.com } 
}
\date{}

\begin{document}
\maketitle
\begin{abstract}
Pre-trained cross-lingual encoders such as mBERT \cite{devlin2019bert} and XLM-R \cite{conneau2019unsupervised} have proven impressively effective at enabling transfer-learning of NLP systems from high-resource languages to low-resource languages. This success comes despite the fact that there is no explicit objective to align the contextual embeddings of words/sentences with similar meanings across languages together in the same space. In this paper, we present a new method for learning multilingual encoders, \xmodel (\textbf{A}ligned \textbf{M}ultilingual \textbf{B}idirectional \textbf{E}ncode\textbf{R}). \xmodel is trained on additional parallel data using two \emph{explicit} alignment objectives that align the multilingual representations at different granularities. We conduct experiments on zero-shot cross-lingual transfer learning for different tasks including sequence tagging, sentence retrieval and sentence classification. Experimental results on the tasks in the XTREME benchmark~\cite{hu2020xtreme} show that \xmodel obtains gains of up to 1.1 average F1 score on sequence tagging and up to 27.3 average accuracy on retrieval over the XLM-R-large model which has 3.2x the parameters of \xmodel. Our code and models are available at \url{http://github.com/junjiehu/amber}.

\end{abstract}

\begin{figure*}[th!]
    \centering
    \includegraphics[width=0.9\textwidth]{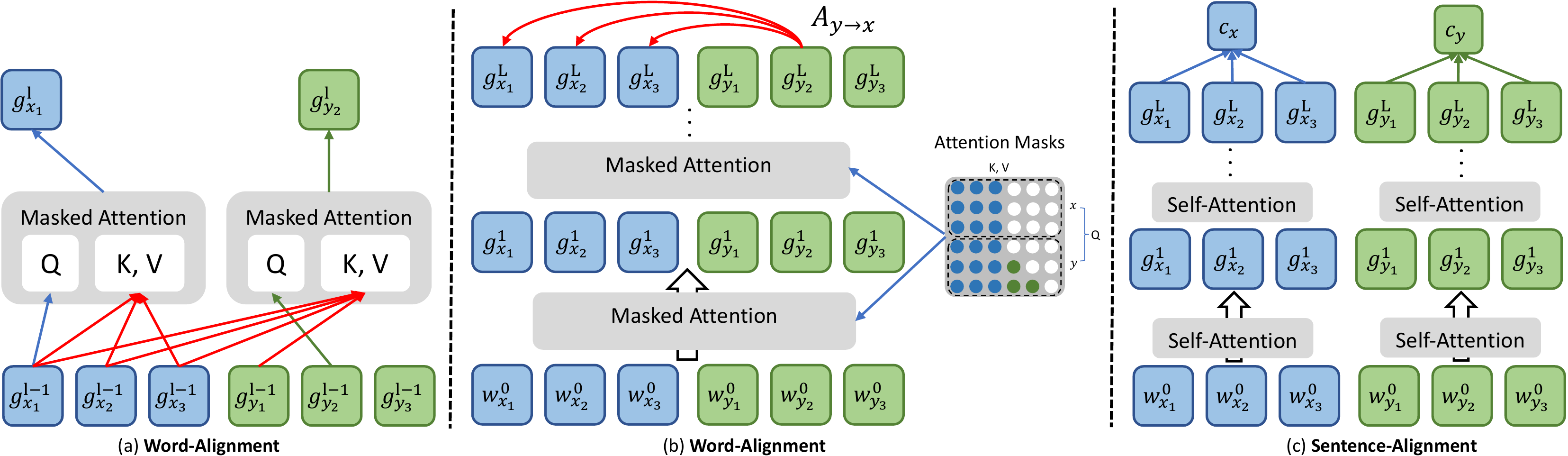}
    \vspace{-0.3cm}
    \caption{(a-b) show the computation of the target-to-source attention matrix used for the \textbf{word alignment} objective: (a) Masked attention for source/target (blue/green) sentences on the $l$-th layer; (b) Attention from $y$ to $x$ on the top layer. (c) shows the separate encoding of source/target sentences for the \textbf{sentence alignment} objective.}
    \label{fig:amber_main}
\end{figure*}

\section{Introduction}
\label{sec:intro}



Cross-lingual embeddings, both traditional non-contextualized word embeddings~\cite{faruqui2014improving} and the more recent contextualized word embeddings \cite{devlin2019bert}, are an essential tool for cross-lingual transfer in downstream applications. In particular, multilingual contextualized word representations have proven effective in reducing the amount of supervision needed in a variety of cross-lingual NLP tasks such as sequence labeling~\cite{pires2019multilingual}, question answering~\cite{artetxe2020cross}, parsing~\cite{wang-etal-2019-cross}, sentence classification~\cite{wu2019beto} and retrieval~\cite{yang2019improving}.

Some attempts at training multilingual representations \citep{devlin2019bert,conneau2019unsupervised} simply train a (masked) language model on monolingual data from many languages. These methods can only \emph{implicitly} learn which words and structures correspond to each-other across languages in an entirely unsupervised fashion, but are nonetheless quite effective empirically \cite{wu2019emerging,K2020Cross-Lingual}.
On the other hand, some methods directly leverage multilingual parallel corpora~\cite{mccann2017learned,eriguchi2018zero,conneau2019cross,huang2019unicoder,siddhant2019evaluating}, which gives some degree of supervision implicitly aligning the words in the two languages.
However, the pressure on the model to learn clear correspondences between the contextualized representations in the two languages is still implicit and somewhat weak.
Because of this, several follow-up works \citep{schuster-etal-2019-cross,wang2020crosslingual,cao2020multilingual} have proposed methods that use word alignments from parallel corpora as the supervision signals to align multilingual contextualized representations, albeit in a \emph{post-hoc} fashion.


In this work, we propose a training regimen for learning contextualized word representations that encourages symmetry at both the word and sentence levels \emph{at training time}.  
Our word-level alignment objective is inspired by work in machine translation that defines objectives encouraging consistency between the source-to-target and target-to-source attention matrices~\cite{cohn2016incorporating}. Our sentence-level alignment objective encourages prediction of the correct translations within a mini-batch for a given source sentence, which is inspired by work on learning multilingual sentence representations~\cite{yang2019improving,wieting-etal-2019-simple}. In experiments, we evaluate the zero-shot cross-lingual transfer performance of \xmodel on four different NLP tasks in the XTREME benchmark~\cite{hu2020xtreme} including part-of-speech (POS) tagging, paraphrase classification, and sentence retrieval. We show that \xmodel obtains gains of up to 1.1 average F1 score on cross-lingual POS tagging, up to 27.3 average accuracy score on sentence retrieval, and achieves competitive accuracy in paraphrase classification when compared with the XLM-R-large model. This is despite the fact that XLM-R-large is trained on data 23.8x as large\footnote{\xmodel is trained on 26GB parallel data and 80GB monolingual Wikipedia data, while XLM-R-large is trained on 2.5TB monolingual CommonCrawl data.} and has 3.2x parameters of \xmodel. This shows that compared to large amounts of monolingual data, even a small amount of parallel data leads to significantly better cross-lingual transfer learning. 



\section{Cross-lingual Alignment}
\label{sec:method}
This section describes three objectives for training contextualized embeddings. We denote the monolingual and parallel data as $\Mcal$ and $\Pcal$ respectively. 


\vspace{0.3cm}
\noindent \textbf{Masked Language Modeling (MLM)}
A masked language modeling objective takes a pair of sentences $x, y$, and optimizes the prediction of randomly masked tokens in the concatenation of the sentence pair as follows:
\begin{align}
    \ell_\text{MLM}(x,y) = - \EE_{s\sim [1,|z|]} \log P(z_{s} | z_{\setminus s}),
\end{align}
where $z$ is the concatenation of the sentence pair $z=[x; y]$, $z_s$ are the masked tokens randomly sampled from $z$, and $z_{\setminus s}$ indicates all the other tokens except the masked ones. 

In the standard monolingual setting, $x,y$ are two contiguous sentences in a monolingual corpus. In \citet{conneau2019cross}, $x,y$ are two sentences in different languages from a parallel corpus, an objective we will refer to as Translation Language Modeling (TLM).


\noindent \textbf{Sentence Alignment}
Our first proposed objective encourages cross-lingual alignment of sentence representations. For a source-target sentence pair $(x,y)$ in the parallel corpus, we separately calculate sentence embeddings denoted as $\cbb_x, \cbb_y$ by averaging the embeddings in the final layer as the sentence embeddings.%
\footnote{In comparison, mBERT encodes a sentence pair jointly, then uses the CLS token embedding to perform its next sentence prediction task.} We then encourage the model to predict the correct translation $y$ given a source sentence $x$. To do so, we model the conditional probability of a candidate sentence $y$ being the correct translation of a source sentence $x$ as:
\begin{align}\label{eq:rank}
    P(y|x) &= {e^{\cbb_x^T\cbb_y} \over \sum_{y'\in \Mcal\cup \Pcal} e^{\cbb_x^T\cbb_{y'}}} .
\end{align}
where $y'$ can be any sentence in any language. Since the normalization term in Eq.~(\ref{eq:rank}) is intractable, we approximate $P(y|x)$ by sampling $y'$ within a mini-batch $\Bcal$ rather than $\Mcal \cup \Pcal$.
We then define the sentence alignment loss as the average negative log-likelihood of the above probability: 
\begin{align}
    \ell_\text{SA}(x,y) = - \log P(y|x).
\end{align}

\noindent \textbf{Bidirectional Word Alignment}
Our second proposed objective encourages alignment of word embeddings by leveraging the attention mechanism in the Transformer model.
Motivated by the work on encouraging the consistency between the source-to-target and target-to-source translations \cite{cohn2016incorporating,he2016dual}, we create two different attention masks as the inputs to the Transformer model, and obtain two attention matrices in the top layer of the Transformer model. We compute the target-to-source attention matrix $\Ab_{y\rightarrow x}$ as follows:

\begin{resizealign} \label{eq:t2s_attn} 
    & \gb^l_{y_i} = \text{Attn}(\Qb=\gb^{l-1}_{y_i}, \Kb\Vb=\gb^{l-1}_{\textcolor{red}{[y_{<i}; x]}}; W^{l}) , \\
    & \gb^l_{x_j} = \text{Attn}(\Qb=\gb_{x_j}^{l-1}, \Kb\Vb=\gb^{l-1}_{\textcolor{red}{x}}; W^{l}) , \\
    & \text{Attn}(\Qb\Kb\Vb; W) = \text{softmax}(\Qb\Wb^q (\Kb \Wb^k)^T) \Vb \Wb^v \\
    & A_{y\rightarrow x}[i,j] = \gb_{y_i}^L \cdot \gb_{x_j}^L .
\end{resizealign}


\noindent where $\gb_{y_t}^{l}$ is the embedding of the $t$-th word in $y$ on the $l$-th layer, $A_{y\rightarrow x}[i,j]$ is the $(i,j)$-th value in the attention matrix from $y$ to $x$, and $W=\{\Wb^q, \Wb^k, \Wb^v\}$ are the linear projection weights for $Q, K, V$ respectively. We compute the source-to-target matrix $\Ab_{x\rightarrow y}$ by switching $x$ and $y$. 
 
To encourage the model to align source and target words in both directions, we aim to minimize the distance between the forward and backward attention matrices. Similarly to \citet{cohn2016incorporating}, we aim to maximize the trace of two attention matrices, i.e., $\tr{({{\Ab_{y\rightarrow x}}}^T \Ab_{x\rightarrow y})}$. Since the attention scores are normalized in $[0,1]$, the trace of two attention matrices is upper bounded by $\min(|x|, |y|)$, and the maximum value is obtained when the two matrices are identical. Since the Transformer generates multiple attention heads, we average the trace of the bidirectional attention matrices generated by all the heads denoted by the superscript $h$
\begin{align} \label{eq:sent_align}
    \ell_\text{WA}(x,y) =  1 - {1\over H}\sum_{h=1}^H{\tr{({\Ab^h_{y\rightarrow x}}^T \Ab^h_{x\rightarrow y})} \over \min(|x|,|y|)}.
\end{align}

Notably, in the target-to-source attention in Eq~(\ref{eq:t2s_attn}), with attention masking we enforce a constraint that the $t$-th token in $y$ can only perform attention over its preceding tokens $y_{<t}$ and the source tokens in $x$. This is particularly useful to control the information access of the query token $y_t$, in a manner similar to that of the decoding stage of NMT. Without attention masking, the standard Transformer performs self-attention over all tokens, i.e., $Q=K=\gb^h_{
z}$, and minimizing the distance between the two attention matrices by Eq.~(\ref{eq:sent_align}) might lead to a trivial solution where $\Wb^q \approx \Wb^k$.

\noindent \textbf{Combined Objective}
Finally we combine the masked language modeling objective with the alignment objectives and obtain the total loss in Eq.~(\ref{eq:total_loss}). Notice that in each iteration, we sample a mini-batch of sentence pairs from $\Mcal\cup\Pcal$. 
\begin{align} \label{eq:total_loss}
    \Lcal =& \EE_{(x,y)\in \Mcal\cup\Pcal}~\ell_\text{MLM}(x,y) \\ \nonumber
           &+\EE_{(x,y)\in \Pcal}~[\ell_\text{SA}(x,y) + \ell_\text{WA}(x,y)]
\end{align}


\begin{table}[]
\setlength{\tabcolsep}{2.0pt}
\resizebox{\columnwidth}{!}{%
\begin{tabular}{l cccccc}
\toprule
Model       & Data &  Langs & Vocab & Layers & Parameters & Ratio\\ \midrule
\xmodel     & Wiki \& MT   & 104   & 120K     & 12  & 172M & 1.0   \\
mBERT       & Wiki     & 104   & 120K     & 12 & 172M & 1.0     \\ 
XLM-15      & Wiki \& MT   & 15    & 95K      & 12  & 250M & 1.5x   \\ 
XLM-100     & Wiki   & 100   & 200K     & 12  & 570M & 3.3x   \\ 
XLM-R-base  & CommonCrawl & 100   & 250K     & 12  & 270M  & 1.6x   \\ 
XLM-R-large & CommonCrawl & 100   & 250K     & 24  & 550M  & 3.2x   \\ 
Unicoder  & CommonCrawl \& MT  & 100   & 250K     & 12  & 270M  & 1.6x   \\ 
\bottomrule
\end{tabular}
}
\vspace{-0.3cm}
\caption{Details of baseline and state-of-the-art models. }
\vspace{-0.4cm}
\label{tab:params}
\end{table}

\section{Experiments}
\subsection{Training setup}
Following the setting of \citet{hu2020xtreme}, we focus on \textit{zero-shot cross-lingual transfer} where we fine-tune models on English annotations and apply the models to predict on non-English data. 

\noindent \textbf{Models}: Table~\ref{tab:params} shows details of models in comparison.  We adopt the same architecture as mBERT for \xmodel. Notably, \xmodel, XLM-15 and Unicoder are trained on the additional parallel data, while the others are trained only on monolingual data. Besides, XLM-R-base/large models have 2.6x/4.8x the parameters of \xmodel and are trained on the larger CommonCrawl corpus. We use a simple setting for our \xmodel variants in the ablation study to show the effectiveness of our proposed alignment objectives without other confounding factors such as model sizes, hyper-parameters and tokenizations in different existing studies. 

\noindent \textbf{Pre-training}: We train \xmodel on the Wikipedia data for 1M steps first using the default hyper-parameters as mBERT\footnote{\url{https://github.com/google-research/bert}} except that we use a larger batch of 8,192 sentence pairs, as this has proven effective in \citet{liu2019roberta}. We then continue training the model by our objectives for another 1M steps with a batch of 2,048 sentence pairs from Wikipedia corpus and parallel corpus which is used to train XLM-15 \cite{conneau2019cross}. We use the same monolingual data as mBERT and follow \citet{conneau2019cross} to prepare the parallel data with one change to maintain truecasing. We set the maximum number of subwords in the concatenation of each sentence pair to 256 and use 10k warmup steps with the peak learning rate of 1e-4 and a linear decay of the learning rate. We train \xmodel on TPU v3 for about 1 week.


\subsection{Datasets}
\noindent \textbf{Cross-lingual Part-Of-Speech (POS)} contains data in 13 languages from the Universal Dependencies v2.3~\cite{nivre2018universal}. 

\noindent \textbf{PAWS-X}~\cite{Yang2019paws-x} is a paraphrase detection dataset. We train on the English data \cite{Zhang2019paws}, and evaluate the prediction accuracy on the test set translated into 4 other languages. 

\noindent \textbf{XNLI}~\cite{conneau2018xnli} is a natural language inference dataset in 15 languages. We train models on the English MultiNLI training data \cite{Williams2018multinli}, and evaluate on the other 14. 

\noindent \textbf{Tatoeba}~\cite{Artetxe2019massively} is a testbed for parallel sentence identification. We select the 14 non-English languages covered by our parallel data, and follow the setup in \citet{hu2020xtreme} finding the English translation for a given a non-English sentence with maximum cosine similarity.


\subsection{Result Analysis}


In Table~\ref{tab:all-results}, we show the average results over all languages in all the tasks, and show detailed results for each language in Appendix~\ref{sec:detailed_results}. First, we find that our re-trained mBERT (\xmodel with MLM) performs better than the publicly available mBERT on all the tasks, confirming the utility of pre-training BERT models with larger batches for more steps~\cite{liu2019roberta}. Second, \xmodel trained by the word alignment objective obtains a comparable average F1 score with respect to the best performing model (Unicoder) in the POS tagging task, which shows the effectiveness of the word-level alignment in the syntactic structure prediction tasks at the token level. Besides, it is worth noting that Unicoder is initialized from the larger XLM-R-base model that is pre-trained on a larger corpus than \xmodel, and Unicoder improves over XLM-R-base on all tasks. Third, for the sentence classification tasks, \xmodel trained with our explicit alignment objectives obtain a larger gain (up to 2.1 average accuracy score in PAWS-X, and 3.9 average accuracy score in XNLI) than \xmodel with only the MLM objective.  Although we find that \xmodel trained with only the MLM objective falls behind existing XLM/XLM-R/Unicoder models with many more parameters, 
\xmodel trained with our alignment objectives significantly narrows the gap of classification accuracy with respect to XLM/XLM-R/Unicoder. Finally, for sentence retrieval tasks, we find that XLM-15 and Unicoder are both trained on additional parallel data, outperforming the other existing models trained only on monolingual data. Using additional parallel data, \xmodel with MLM and TLM objectives also significantly improves over \xmodel with the MLM objective by 15.6 average accuracy score, while combining our word-level alignment objective yields a marginal improvement over \xmodel with MLM and TLM objectives. However, adding the sentence-level alignment objective, \xmodel trained by the combined objective can further improve \xmodel with the MLM and word-level alignment objectives by 19.1 average accuracy score. This confirms our intuition that the explicit sentence-level objective can effectively leverage the alignment supervision in the parallel corpus, and encourage contextualized sentence representations of aligned pairs to be close according to the cosine similarity metric.

\begin{table}[]
\centering
\setlength{\tabcolsep}{2.0pt}
\resizebox{0.48\textwidth}{!}{%
\begin{tabular}{l|cccc}
\toprule
Model               & POS           & PAWS-X        & XNLI          & Tatoeba       \\ \midrule
mBERT (public)      & 68.5          & 86.2          & 65.4          & 45.6          \\ 
XLM-15              & 68.8          & 88.0          & 72.6          & \textbf{77.2} \\ 
XLM-100             & 69.5          & 86.4          & 69.1          & 36.6          \\ 
XLM-R-base          & 68.8          & 87.4          & 73.4          & 57.6          \\
XLM-R-large         & 70.0 & \textbf{89.4} & \textbf{79.2} & 60.6          \\  
Unicoder            & \textbf{71.7} & 88.1          & 74.8          & 72.2 \\ \midrule
\xmodel(MLM)           & 69.8          & 87.1          & 67.7          & 52.6          \\ 
\xmodel(MLM+TLM)       & 70.5          & 87.7          & 70.9          & 68.2          \\ 
\xmodel(MLM+TLM+WA)    & \textbf{71.1} & 89.0          & 71.3          & 68.8          \\ 
\xmodel(MLM+TLM+WA+SA) & 70.5          & \textbf{89.2} & \textbf{71.6} & \textbf{87.9} \\ \bottomrule
\end{tabular}%
\vspace{-3mm}
}
\caption{Overall results on POS, PAWS-X, XNLI, Tatoeba tasks. Bold numbers highlight the highest scores across languages on the existing models (upper part) and \xmodel variants (bottom part).}
\vspace{-5mm}
\label{tab:all-results}
\end{table}

\subsection{How does alignment help by language?}
\begin{figure}
    \centering
    \includegraphics[width=\linewidth]{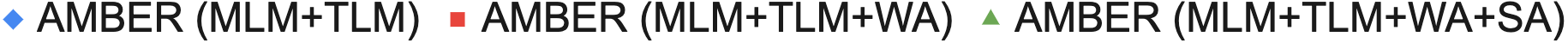}
    \begin{subfigure}{0.238\textwidth}
        \includegraphics[width=\linewidth]{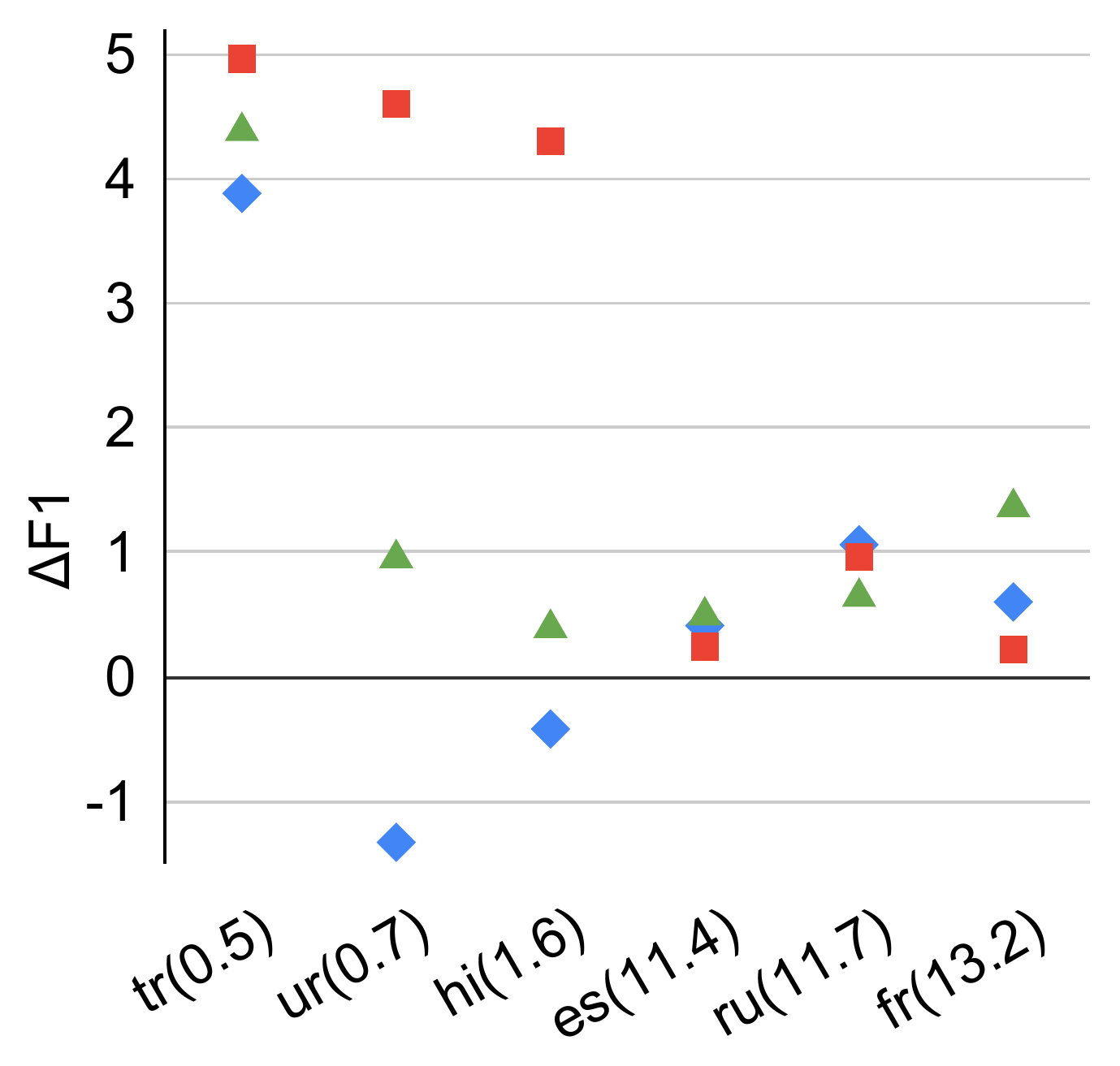}
          \caption{Part-of-Speech}
          \label{fig:pos}
      \end{subfigure}
    \begin{subfigure}{0.237\textwidth}
        \includegraphics[width=\linewidth]{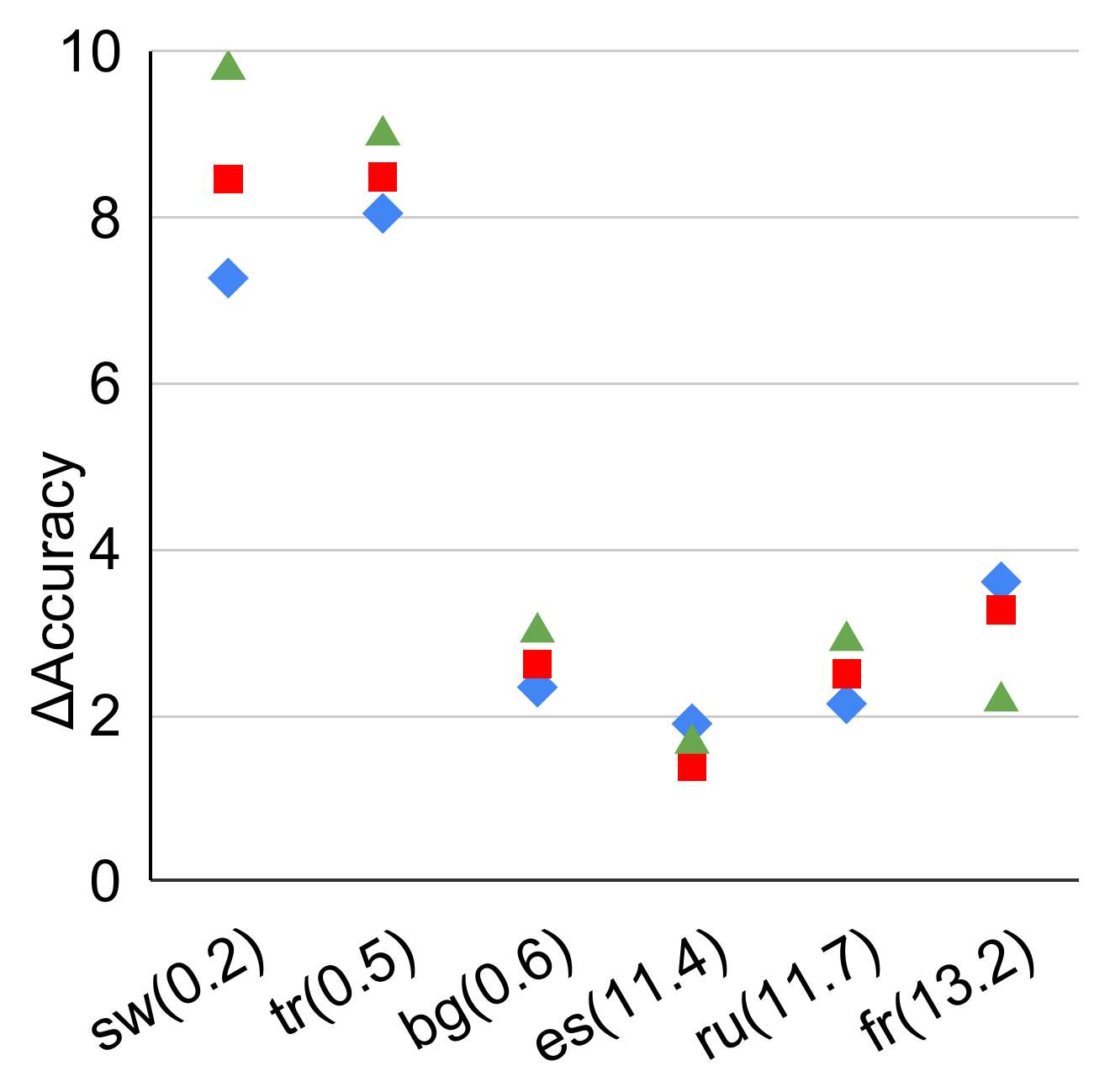}
          \caption{XNLI}
          \label{fig:xnli}
      \end{subfigure}
      \vspace{-0.35cm}
    \caption{Performance difference between \xmodel trained with alignments on parallel data and \xmodel (MLM). Languages are sorted by no.~of parallel data (Million) used for training \xmodel with alignments.}
    \label{fig:lang}
    \vspace{-0.4cm}
\end{figure}

In Figure~\ref{fig:lang}, we investigate the improvement of the alignment objectives over the MLM objective on low-resourced and high-resourced languages, by computing the performance difference between \xmodel trained with alignment objectives and \xmodel (MLM). First, we find that \xmodel trained with alignment objectives significantly improves the performance on languages with relatively small amounts of parallel data, such as Turkish, Urdu, Swahili, while the improvement on high-resourced languages is marginal. Through a further analysis (Appendix~\ref{sec:detailed_results}), we observe that \xmodel (MLM) performs worse on these low-resourced and morphologically rich languages than on high-resourced Indo-European languages, while \xmodel trained with alignment objectives can effectively bridge the gap. Moreover, \xmodel trained with our word-level alignment objective yields the highest improvement on these low-resourced languages on the POS task, and \xmodel trained with sentence-level alignment performs the best on XNLI.

\subsection{Alignment with Attention vs Dictionary }
Recent studies~\cite{cao2020multilingual,wang2020crosslingual} have proposed to use a bilingual dictionary to align cross-lingual word representations. Compared with these methods, our word-level alignment objective encourages the model to automatically discover word alignment patterns from the parallel corpus in an end-to-end training process, which avoids potential errors accumulated in separate steps of the pipeline. Furthermore, an existing dictionary may not have all the translations for source words, especially for words with multiple senses. Even if the dictionary is relatively complete, it also requires a heuristic way to find the corresponding substrings in the parallel sentences for alignment. If we use a word alignment tool to extract a bilingual dictionary in a pipeline, errors may accumulate, hurting the accuracy of the model. Besides, \citet{wang2020crosslingual} is limited in aligning only fixed contextual embeddings from the model’s top layer. Finally, we also compare \xmodel trained with all the objectives and \citet{cao2020multilingual} on a subset of languages on XNLI in Table~\ref{tab:cao_xnli}. We find that our full model obtains a gain of 1.8 average F1 score.

\begin{table}[]
\centering
\setlength{\tabcolsep}{2.0pt}
\resizebox{0.48\textwidth}{!}{%
    \begin{tabular}{l|cccccc|c} \toprule
         Methods & en & bg & de & el & es & fr & Avg. \\ \midrule
         \citet{cao2020multilingual} & 80.1 & 73.4 & 73.1 & 71.4 & 75.5 & 74.5 & 74.7  \\
         \xmodel (full)       & 84.7 & 74.3 & 74.2 & 72.5 & 76.9 & 76.6 & 76.5 \\ \bottomrule
    \end{tabular}}
    \caption{F1 scores of \xmodel trained with all objectives and \citet{cao2020multilingual} on 6 languages on XNLI.}
    \label{tab:cao_xnli}
\end{table}

\section{Related Work}
\label{sec:related}
While cross-lingual alignment is a long-standing challenge dating back to the early stage of research in word alignment~\cite{brown-etal-1993-mathematics}, cross-lingual embeddings~\cite{faruqui2014improving,xing2015normalized,devlin2019bert,conneau2019unsupervised} are highly promising in their easy integration into neural network models for a variety of cross-lingual applications. Analysis studies on recent cross-lingual contextualized representations~\cite{pires2019multilingual,wu2019beto,hu2020xtreme,siddhant2019evaluating} further demonstrates this advantage for zero-shot cross-lingual transfer in a representative set of languages and tasks. In particular to improve cross-lingual transfer, some attempts directly leverage multilingual parallel corpus to train contextualized representations~\cite{mccann2017learned,eriguchi2018zero,conneau2019cross,huang2019unicoder} with the hope of aligning words implicitly. The other line of work uses word alignments from parallel corpora as the alignment supervision in a post-hoc fashion~\cite{cao2020multilingual,wang2020crosslingual}. Notably, \xmodel does not rely on any word alignment tools, and explicitly encourage the correspondence both on the word and sentence level.

\section{Discussion and Future Work}
In this paper, we demonstrate the effectiveness of our proposed explicit alignment objectives in learning better cross-lingual representations for downstream tasks. Nonetheless, several challenging and promising directions can be considered in the future. First, most existing multilingual models tokenize words into subword units, which makes the alignment less interpretable. How to align a span of subword units with meaningful semantics at the phrase level deserves further investigation. Second, several studies~\cite{ghader-monz-2017-attention,li-etal-2019-word} have shown that attention may fail to capture word alignment for some language pairs, and a few works~\cite{legrand-etal-2016-neural,alkhouli-etal-2018-alignment} proposed neural word alignment to improve the word alignment quality. Incorporating such recent advances into the alignment objective is one future direction. Third, how to fine-tune a well-aligned multilingual model on English annotations without catastrophic forgetting of the alignment information is a potential way to improve cross-lingual generalization on the downstream applications.


\section*{Acknowledgements}
We’d like to thank Yinfei Yang and Wei-Cheng Chang for answering our questions on the data and code. JH and GN were supported in part by a Google Faculty Award, and NSF Award \#1761548.

\bibliography{ref.bib}
\bibliographystyle{acl_natbib}

\clearpage
\appendix
\input{appendix}

\end{document}

%% file: appendix.tex
\section{Appendices}
\label{sec:appendix}

\subsection{Training Details for Reproducibility} \label{sec:details}
Although English is not the best source language for some target languages \cite{Lin2019transfer_languages}, this zero-shot cross-lingual transfer setting is still practical useful as many NLP tasks only have English annotations. In the following paragraphs, we show details for reproducing our results on zero-shot cross-lingual transfer setting.

\paragraph{\textbf{Model:}} We use the same architecture as mBERT for \xmodel, and we build our \xmodel trained with the alignment objectives on top of the original mBERT implementation at \url{https://github.com/google-research/bert}, and are released at \url{http://github.com/junjiehu/amber}.

\paragraph{\textbf{Pre-training:}} We first train the model on the Wikipedia data for 1M steps using the default hyper-parameters in the original repository except that we use a larger batch of 8,192 sentence pairs. The max number of subwords in the concatenation of each sentence pair is set to 256. To continue training \xmodel with additional objectives on parallel data, we use 10K warmup steps with the peak learning rate of 1e-4, and use a linear decay of the learning rate. All models are pre-trained with our proposed objectives on TPU v3, and we use the same hyper-parameter setting for our \xmodel variants in the experiments. We follow the practice of mBERT at \url{https://github.com/google-research/bert/blob/master/multilingual.md#data-source-and-sampling} to sample from mutlilingual data for training. We select the checkpoint of all models at the 1M step for a fair comparison. It takes about 1 week to finish the pre-training.

\paragraph{\textbf{Fine-tuning:}} For fine-tuning the models on the downstream applications, we use the constant learning rate of 2e-5 as suggested in the original paper~\cite{devlin2019bert}. We fine-tune all the models for 10 epochs on the cross-lingual POS tag prediction task, and 5 epochs on the sentence classification task. We use the batch size of 32 for all the models. All models are fine-tuned on 2080Ti GPUs, and the training can be finished within 1 day.

\paragraph{\textbf{Datasets:}} We use the same parallel data that is used to train XLM-15. The parallel data can be processed by this script: \url{https://github.com/facebookresearch/XLM/blob/master/get-data-para.sh}. All the datasets in the downstream applications can be downloaded by the script at \url{https://github.com/google-research/xtreme/blob/master/scripts/download_data.sh}. Table~\ref{tab:languages} lists all the data statistic of parallel data by languages.

\begin{table*}[]
\centering
\resizebox{\textwidth}{!}{%
\begin{tabular}{l c c c l l ccccc}
\toprule
Language & \begin{tabular}[c]{@{}l@{}}ISO\\ 639-1\\ code\end{tabular} &
\begin{tabular}[c]{@{}l@{}}\# Parallel\\ sentences (in\\ millions)\end{tabular} &
\begin{tabular}[c]{@{}l@{}}\# Wikipedia\\ articles (in\\ millions)\end{tabular} & Script & \begin{tabular}[c]{@{}l@{}}Language\\ family\end{tabular} & \begin{tabular}[c]{@{}l@{}}Diacritics /\\ special\\ characters\end{tabular} & \begin{tabular}[c]{@{}l@{}}Extensive\\ compound-\\ ing\end{tabular} & \begin{tabular}[c]{@{}l@{}}Bound\\ words /\\ clitics\end{tabular} & \begin{tabular}[c]{@{}l@{}}Inflec-\\ tion\end{tabular} & \begin{tabular}[c]{@{}l@{}}Deriva-\\ tion\end{tabular} \\ \midrule
Arabic & ar & 9.8 & 1.02 & Arabic & Afro-Asiatic & X &  & X & X &   \\
Bulgarian & bg & 0.6 & 0.26 & Cyrillic & IE: Slavic & X &  & X & X &   \\
English & en & 40.2 & 5.98 & Latin & IE: Germanic &  &  &  &  &   \\
French & fr & 13.2 & 2.16 & Latin & IE: Romance & X &  & X &  &   \\
German & de & 9.3 & 2.37 & Latin & IE: Germanic &  & X &  & X &   \\
Greek & el & 4.0 & 0.17 & Greek & IE: Greek & X & X &  & X &   \\
Hindi & hi & 1.6 &  0.13 & Devanagari & IE: Indo-Aryan & X & X & X & X & X  \\
Mandarin & zh & 9.6 & 1.09 & Chinese ideograms & Sino-Tibetan &  & X &  &  &  \\
Russian & ru & 11.7 & 1.58 & Cyrillic & IE: Slavic &  &  &  & X &   \\
Spanish & es & 11.4 & 1.56 & Latin & IE: Romance & X &  & X &  &   \\
Swahili & sw & 0.2 & 0.05 & Latin & Niger-Congo &  &  & X & X & X  \\
Thai & th & 3.3 & 0.13 & Brahmic & Kra-Dai & X &  &  &  &   \\
Turkish & tr & 0.5 &  0.34 & Latin & Turkic & X & X &  & X & X  \\
Urdu & ur & 0.7 & 0.15 & Perso-Arabic & IE: Indo-Aryan & X & X & X & X & X \\
Vietnamese & vi & 3.5 &  1.24 & Latin & Austro-Asiatic & X &  &  &  &   \\
\bottomrule
\end{tabular}%
}
\caption{Statistics about languages used for pre-training with our alignment objectives. Languages belong to 7 language families, with Indo-European (IE) having the most members. Diacritics / special characters: Language adds diacritics (additional symbols to letters). Compounding: Language makes extensive use of word compounds. Bound words / clitics: Function words attach to other words. Inflection: Words are inflected to represent grammatical meaning (e.g.~case marking). Derivation: A single token can represent entire phrases or sentences.}
\label{tab:languages}
\end{table*}

\subsection{Source-to-target attention matrix} \label{sec:s2t_attention}
We derive the source-to-target attention matrix as follow:

\begin{resizealign} \label{eq:s2t_attn}
    & \gb^l_{x_j} = \text{Attn}(\Qb=\gb^{l-1}_{x_j}, \Kb\Vb=\gb^{l-1}_{\textcolor{red}{[x_{<j}; y]}}; W^{l}) , \\
    & \gb^l_{y_j} = \text{Attn}(\Qb=\gb_{y_i}^{l-1}, \Kb\Vb=\gb^{l-1}_{\textcolor{red}{y}}; W^{l}) , \\
    & \text{Attn}(\Qb\Kb\Vb; W) = \text{softmax}(\Qb\Wb^q (\Kb \Wb^k)^T) \Vb \Wb^v \\
    & A_{x\rightarrow y}[j,i] = \gb_{x_j}^L \cdot \gb_{y_i}^L .
\end{resizealign}

\subsection{Detailed Results} \label{sec:detailed_results}
We show the detailed results over all languages on the cross-lingual POS task in Table~\ref{tab:conll17-align}, on the PAWS-X task in Table~\ref{tab:pawsx-align}, on the XNLI task in Table~\ref{tab:xnli-result}, and on the Tatoeba retrieval task in Table~\ref{tab:tatoeba-align}.

\begin{table*}[]
\centering
\resizebox{0.6\textwidth}{!}{%
\begin{tabular}{l|lllll|l}
\toprule
Model           & de            & en            & es            & fr            & zh            & Avg           \\ \midrule
mBERT (public)     & 85.7          & 94.0          & 87.4          & 87.0          & 77.0          & 86.2               \\
XLM-15             & 88.5          & \textbf{94.7} & 89.3          & 89.6          & 78.1          & 88.0               \\
XLM-100            & 85.9          & 94.0          & 88.3          & 87.4          & 76.5          & 86.4               \\
XLM-R-base          & 87.0          & 94.2          & 88.6          & 88.7          & 78.5          & 87.4               \\
XLM-R-large         & \textbf{89.7} & \textbf{94.7} & \textbf{90.1} & \textbf{90.4} & \textbf{82.3} & \textbf{89.4}      \\ \midrule
\xmodel (MLM, our mBERT)    & 87.3          & 93.9          & 87.5          & 87.8          & 78.8          & 87.1               \\
\xmodel (MLM+TLM)                & 87.6          & \textbf{95.8} & 87.4          & 88.9          & 78.7          & 87.7               \\
\xmodel(MLM+TLM+WA) & 88.9 & 95.5          & 88.9 & \textbf{90.7} & \textbf{81.1} & 89.0      \\
\xmodel(MLM+TLM+WA+SA)   & \textbf{89.4}          & 95.6          & \textbf{89.2}          & \textbf{90.7}          & 80.9          & \textbf{89.2} \\\bottomrule         
\end{tabular}%
}
\caption{Accuracy of zero-shot cross-lingual classification on PAWS-X. Bold numbers highlight the highest scores across languages on the existing models (upper part) and \xmodel variants (bottom part).}
\label{tab:pawsx-align}
\end{table*}

\begin{table*}[!t]
\centering
\resizebox{\textwidth}{!}{%
\begin{tabular}{l|lllllllllllll|l}
\toprule
models       & ar            & bg            & de            & el            & en            & es            & fr            & hi            & ru            & tr            & ur            & vi            & zh            & Avg  \\ \midrule
mBERT (public)         & 14.9          & 85.2          & 89.3          & 82.8          & 95.3          & 85.7          & 84.1          & 65.1          & 86.0          & 67.5          & 57.4          & 18.5          & 58.9          & 68.5 \\ 
XLM-15       & 17.5          & 86.1          & 89.3          & 85.4          & 95.7          & 85.9          & 84.9          & 63.9          & 86.8          & 69.3          & 55.1          & 18.0          & 57.2          & 68.8 \\ 
XLM-100      & 17.1          & 85.8          & 89.3          & 85.7          & 95.4          & 85.3          & 84.3          & 67.0          & 87.1          & 65.0          & 62.0          & 19.2          & \textbf{60.2}          & 69.5 \\ 
XLM-R-base    & 17.6          & \textbf{88.5} & 91.1          & \textbf{88.2} & 95.8          & 87.2          & 85.7          & 70.1          & 88.9          & 72.7          & 61.6          & 19.2          & 27.9          & 68.8 \\ 
XLM-R-large   & \textbf{18.1} & 87.4          & \textbf{91.9} & 87.9          & \textbf{96.3} & \textbf{87.8} & \textbf{87.3} & \textbf{76.1} & \textbf{89.9} & \textbf{74.3} & \textbf{67.6} & \textbf{19.5} & 26.5          & \textbf{70.0} \\ \midrule
\xmodel (MLM, our mBERT) & 15.4          & 86.6          & 90.1          & 84.3          & 95.5          & 86.5          & 84.6          & 68.2          & 86.8          & 69.0          & 59.2          & 18.7          & 62.1          & 69.8 \\ 
\xmodel (MLM+TLM)          & \textbf{16.0} & \textbf{87.2} & \textbf{91.5} & \textbf{86.4} & \textbf{95.7} & 86.9          & 85.2          & 67.7          & \textbf{87.9} & 72.9          & 57.9          & 19.1          & \textbf{62.7} & 70.5 \\ 
\xmodel (MLM+TLM+WA)       & 14.8          & 86.9          & 90.4          & 84.9          & 95.6          & 86.7          & 84.8          & \textbf{72.5} & 87.8          & \textbf{73.9} & \textbf{63.8} & \textbf{19.5} & 62.3          & \textbf{71.1} \\ 
\xmodel (MLM+TLM+WA+SA)    & 14.6          & 87.1          & 90.6          & 85.9          & 95.5          & \textbf{87.0} & \textbf{86.0} & 68.6          & 87.4          & 73.4          & 60.2          & 18.8          & 61.8          & 70.5 \\ \bottomrule
\end{tabular}}
\caption{F1 scores of part-of-speech tag predictions from the Universal Dependency v2.3. Bold numbers highlight the highest scores across languages on the existing models (upper part) and \xmodel variants (bottom part). }
\label{tab:conll17-align}
\end{table*}

\begin{table*}[h]
\centering
\resizebox{\textwidth}{!}{%
\begin{tabular}{l|lllllllllllllll|l}
\toprule
Models                                  & en            & zh            & es            & de            & ar            & ur            & ru            & bg            & el            & fr            & hi            & sw            & th                        & tr            & vi            & avg           \\ \midrule
mBERT (public)                          & 80.8          & 67.8          & 73.5          & 70.0          & 64.3          & 57.2          & 67.8          & 68.0          & 65.3          & 73.4          & 58.9          & 49.7          & 54.1                      & 60.9          & 69.3          & 65.4          \\ 
XLM-15                                  & 84.1          & 68.8          & 77.8          & 75.7          & 70.4          & 62.2          & 75.0          & 75.7          & 73.3          & 78.0          & 67.3          & 67.5          & 70.5 & 70.0          & 73.0          & 72.6          \\ 
XLM-100                                 & 82.8          & 70.2          & 75.5          & 72.7          & 66.0          & 59.8          & 69.9          & 71.9          & 70.4          & 74.3          & 62.5          & 58.1          & 65.5 & 66.4          & 70.7          & 69.1          \\ 
XLM-R-base                               & 83.9          & 73.6          & 78.3          & 75.2          & 71.9          & 65.4          & 75.1          & 76.7          & 75.4          & 77.4          & 69.1          & 62.2          & 72.0                      & 70.9          & 74.0          & 73.4          \\ 
XLM-R-large                              & \textbf{88.7} & \textbf{78.2} & \textbf{83.7} & \textbf{82.5} & \textbf{77.2} & \textbf{71.7} & \textbf{79.1} & \textbf{83.0} & \textbf{80.8} & \textbf{82.2} & \textbf{75.6} & \textbf{71.2} & \textbf{77.4}             & \textbf{78.0} & \textbf{79.3} & \textbf{79.2} \\ \midrule
\xmodel (MLM, our mBERT) & 82.1          & 71.0          & 75.3          & 72.7          & 66.2          & 60.1          & 70.4          & 71.3          & 67.9          & 74.4          & 63.6          & 50.1          & 55.0                      & 64.2          & 71.6          & 67.7          \\ 
\xmodel (MLM+TLM)        & 84.3          & 71.6          & \textbf{77.2} & 73.9          & 69.1          & 59.6          & 72.5          & 73.6          & 70.9          & \textbf{78.0} & 64.7          & 57.4          & 65.0                      & 72.2          & 73.1          & 70.9          \\
\xmodel (MLM+TLM+WA)     & 84.1          & \textbf{72.1} & 76.6          & \textbf{74.7} & 69.3          & \textbf{61.5} & 72.9          & 73.9          & 71.6          & 77.7          & 65.7          & 58.6          & 65.3                      & 72.7          & \textbf{73.4} & 71.3          \\ 
\xmodel (MLM+TLM+WA+SA)  & \textbf{84.7} & 71.6          & 76.9          & 74.2          & \textbf{70.2} & 61.0          & \textbf{73.3} & \textbf{74.3} & \textbf{72.5} & 76.6          & \textbf{66.2} & \textbf{59.9} & \textbf{65.7}             & \textbf{73.2} & \textbf{73.4} & \textbf{71.6} \\ \bottomrule
\end{tabular} }
\caption{Accuracy of zero-shot crosslingual classification on the XNLI dataset. Bold numbers highlight the highest scores across languages on the existing models (upper part) and \xmodel variants (bottom part).}
\label{tab:xnli-result}
\end{table*}

\begin{table*}[!htbp]
\centering
\resizebox{\textwidth}{!}{%
\begin{tabular}{l|llllllllllllll|l}
\toprule
Method                             & ar            & bg            & de            & el            & es            & fr            & hi            & ru            & sw            & th            & tr            & ur            & vi            & zh            & Avg           \\ \midrule
mBERT (public)                               & 25.8          & 49.3          & 77.2          & 29.8          & 68.7          & 66.3          & 34.8          & 61.2          & 11.5          & 13.7          & 34.8          & 31.6          & 62.0          & 71.6          & 45.6          \\
XLM-15 & \textbf{63.5}	& 71.5	& \textbf{92.6}	& \textbf{73.1}	& \textbf{85.5}	& \textbf{82.5}	& \textbf{81.0} & \textbf{82.0} & 	\textbf{47.9} &	\textbf{90.3} &	\textbf{67.6}	 & \textbf{68.4} &	\textbf{91.1} & 	\textbf{84.1}	& \textbf{77.2} \\
XLM-100                            & 18.2          & 40.0          & 66.2          & 25.6          & 58.4          & 54.5          & 26.5          & 44.8          & 12.6          & 31.8          & 26.2          & 18.1          & 47.1          & 42.2          & 36.6          \\
XLM-R-base                          & 36.8          & 67.6          & 89.9 & 53.7          & 74.0          & 74.1 & 54.2          & 72.5          & 19.0          & 38.3 & 61.1          & 36.6 & 68.4          & 60.8          & 57.6          \\
XLM-R-large                         & 47.5 & \textbf{71.6} & 88.8          & 61.8 & 75.7 & 73.7          & 72.2 & 74.1 & 20.3 & 29.4          & 65.7 & 24.3          & 74.7 & 68.3 & 60.6 \\ \midrule
\xmodel (MLM, our mBERT)                               & 30.7          & 54.9          & 81.4          & 37.7          & 72.7          & 72.7          & 47.5          & 67.5          & 15.1          & 25.7          & 48.3          & 42.6          & 64.6          & 75.1          & 52.6          \\
\xmodel (MLM+TLM)       & 47.1          & 61.8          & 89.0          & 53.8          & 76.3          & 77.9          & 72.3          & 69.8          & 20.5          & 83.4          & 88.1          & 50.0          & 86.9          & 78.0          & 68.2          \\
\xmodel (MLM+TLM+WA)    & 46.8          & 63.3          & 88.8          & 52.2          & 78.3          & 79.5          & 66.9          & 71.6          & 27.4          & 77.2          & 86.9          & 56.5          & 86.5          & 81.6          & 68.8          \\
\xmodel (MLM+TLM+WA+SA) & \textbf{78.5} & \textbf{87.1} & \textbf{95.5} & \textbf{75.3} & \textbf{93.3} & \textbf{92.2} & \textbf{95.0} & \textbf{91.5} & \textbf{52.8} & \textbf{94.5} & \textbf{98.4} & \textbf{84.5} & \textbf{97.4} & \textbf{94.3} & \textbf{87.9} \\ \bottomrule
\end{tabular}%
}
\caption{Sentence retrieval accuracy on the Tatoeba dataset. Bold numbers highlight the highest scores across languages on the existing models (upper part) and \xmodel variants (bottom part).}
\label{tab:tatoeba-align}
\end{table*}

\subsection{Detailed Results on Performance Difference by Languages}
\begin{figure*}
    \centering
    \includegraphics[width=0.8\linewidth]{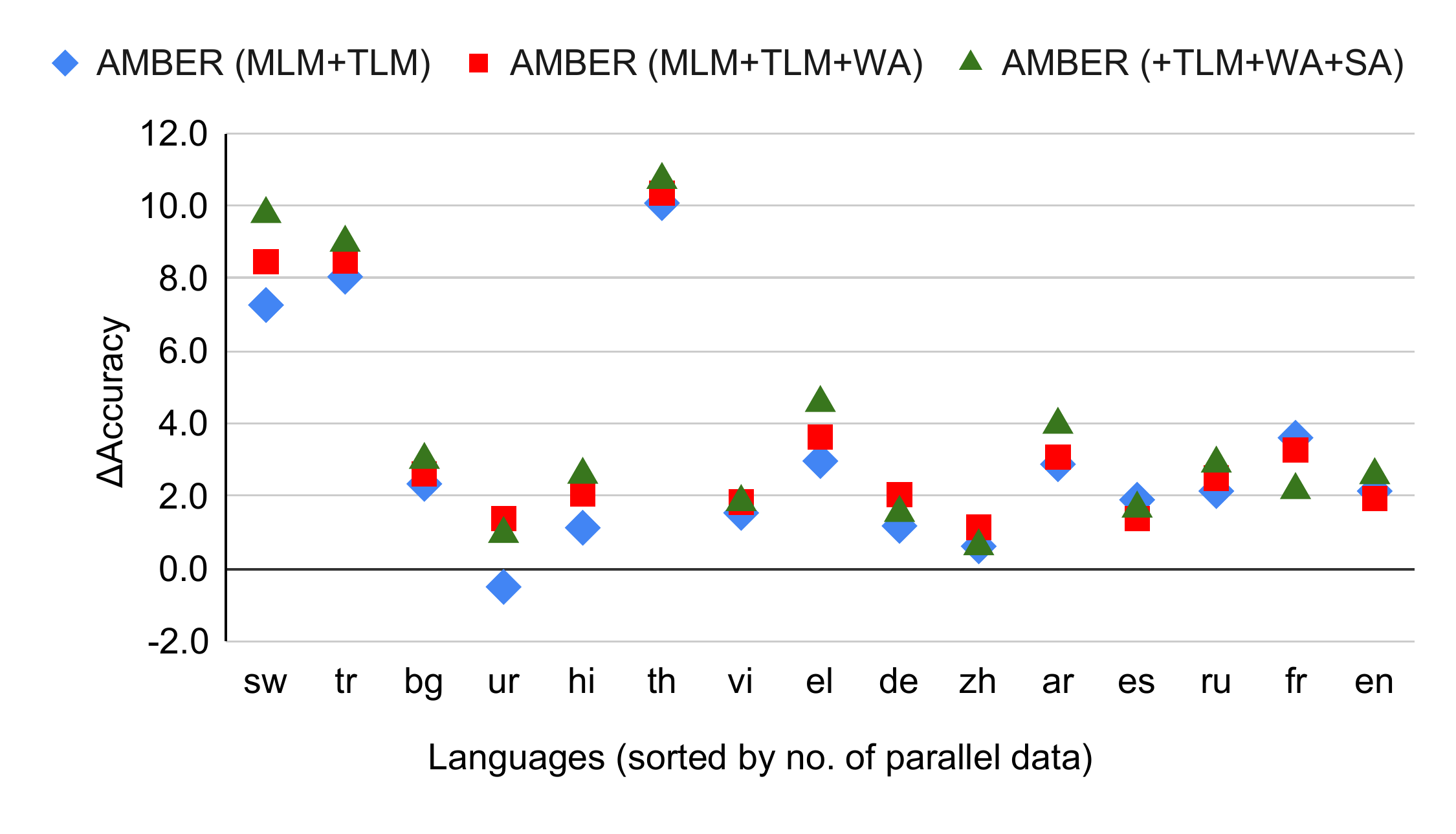}
    \caption{Performance difference between \xmodel trained with alignments on parallel data and \xmodel (MLM) on XNLI task. Languages are sorted by no. of parallel data used for training \xmodel with alignments.}
    \label{fig:xnli_all}
\end{figure*}

\begin{figure*}
    \centering
    \includegraphics[width=0.8\linewidth]{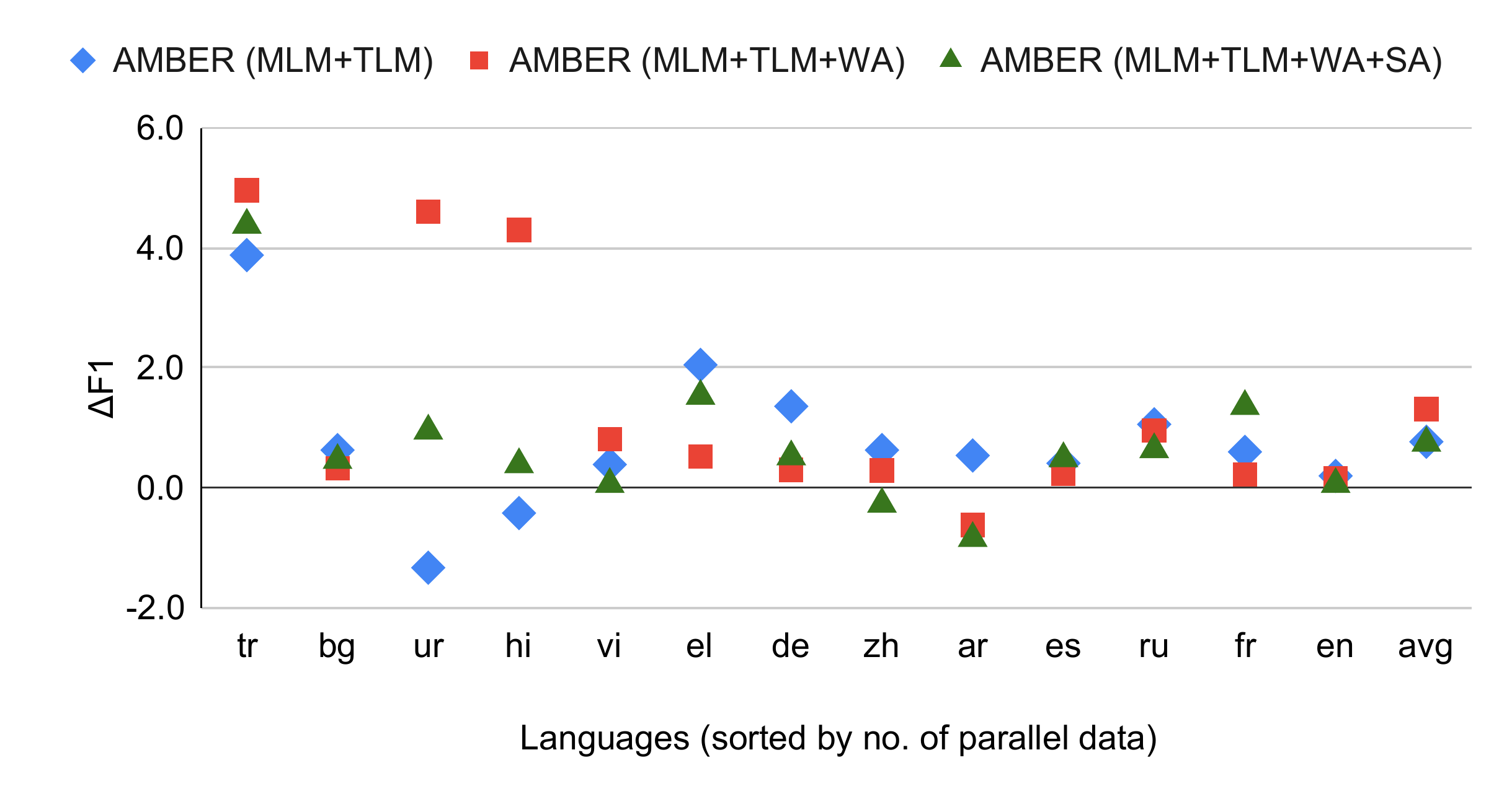}
    \caption{Performance difference between \xmodel trained with alignments on parallel data and \xmodel (MLM) on POS task. Languages are sorted by no. of parallel data used for training \xmodel with alignments.}
    \label{fig:pos_all}
\end{figure*}

Figure~\ref{fig:pos_all} and Figure~\ref{fig:xnli_all} show the performance deference between \xmodel trained with alignment objectives and \xmodel trained with only MLM objective on the POS and XNLI tasks over all languages.

